\documentclass[letterpaper, 10 pt, conference]{ieeeconf}

\IEEEoverridecommandlockouts                            

\usepackage{xcolor}
\usepackage{subcaption}
\usepackage{booktabs,xcolor}
\usepackage{cleveref}
\usepackage{fancyvrb}

\usepackage{graphicx}
\let\labelindent\relax

\usepackage{enumitem}

\setitemize{label=\textbullet, leftmargin=*, nolistsep}
\setenumerate{leftmargin=*, nolistsep}

\usepackage{mdframed}
    \mdfsetup{skipabove=1.5ex,skipbelow=1.5ex}
    \mdfdefinestyle{spacedframe}{%
        linewidth=1pt
    }

\newcommand{\pseudosection}[1]{\vspace{1.2ex}\noindent\textit{#1.~}}

\definecolor{MainPromptFrame}{RGB}{44, 82, 130}
\definecolor{MainPromptBack}{RGB}{235, 242, 250}

\definecolor{StructuralFrame}{RGB}{52, 122, 116}
\definecolor{StructuralBack}{RGB}{235, 247, 245}

\definecolor{ExampleFrame}{RGB}{120, 120, 120}
\definecolor{ExampleBack}{RGB}{245, 245, 245}
\title{\LARGE \bf A Framework for Increasing Situational Awareness In High-Risk HRI}
\title{\LARGE \bf Increasing Situational Awareness In High-Risk Situations}
\title{\LARGE \bf Autonomous VR-Based Risk Annotation \\ for Situational Awareness in Dangerous Settings}
\title{\LARGE \bf Autonomous VR-Based Risk Detection\\for Situational Awareness in Dangerous Settings}

\author{Mohammad Eskandari$^{1}$, Murali Indukuri$^{1}$, Stephanie Lukin$^{2}$, Cynthia Matuszek$^{1}$ 
\thanks{* This work was supported in part by NSF Grants IIS-2024878 and IIS-2145642, and this material is also based on research that is in part supported by the Army Research Laboratory, Grant No. W911NF2120076.}
\thanks{$^{1}$Interactive Robotics and Language Lab, University of Maryland Baltimore County, Computer Science and Electrical Engineering Department, Baltimore, MD, USA.
    {\tt\footnotesize (eskandari|muralii1|cmat)@umbc.edu}}%
\thanks{\textbf{Accepted to RO-MAN 2026}.}
\thanks{$^{2}$DEVCOM Army Research Laboratory, Adelphi, MD, USA.
    {\tt\footnotesize stephanie.m.lukin.civ@army.mil}}%
    }

\begin{document}
\maketitle
\thispagestyle{empty}
\pagestyle{empty}

\begin{abstract}
In high-risk environments such as disaster response, situational awareness depends not only on detecting hazards but also on communicating them clearly to human operators. Vision–Language Models (VLMs) have shown strong potential for scene understanding in safety-critical settings, yet their value as part of human-facing robotic systems remains underexplored. We present a VR-based Human–Robot Interaction framework for studying how VLM-assisted robots can support situational awareness in simulated hazardous environments. In our system, a robot explores a virtual scene and queries a VLM to identify potential hazards and annotate user-facing points of interest. These annotations are presented to a human operator through an immersive VR interface. This framework enables controlled evaluation of both robotic hazard identification and the communication of safety-critical information to users. Results from our study indicate that the annotated VR interface was preferred over the unannotated baseline and that participants reported high clarity, usefulness, and comfort when interacting with the system. These findings suggest that combining VLM-based robotic perception with immersive visualization is a promising approach for supporting situational awareness in hazardous settings.
\end{abstract}
\begin{figure}[htbp]
    \centering
    \includegraphics[width=0.7\linewidth]{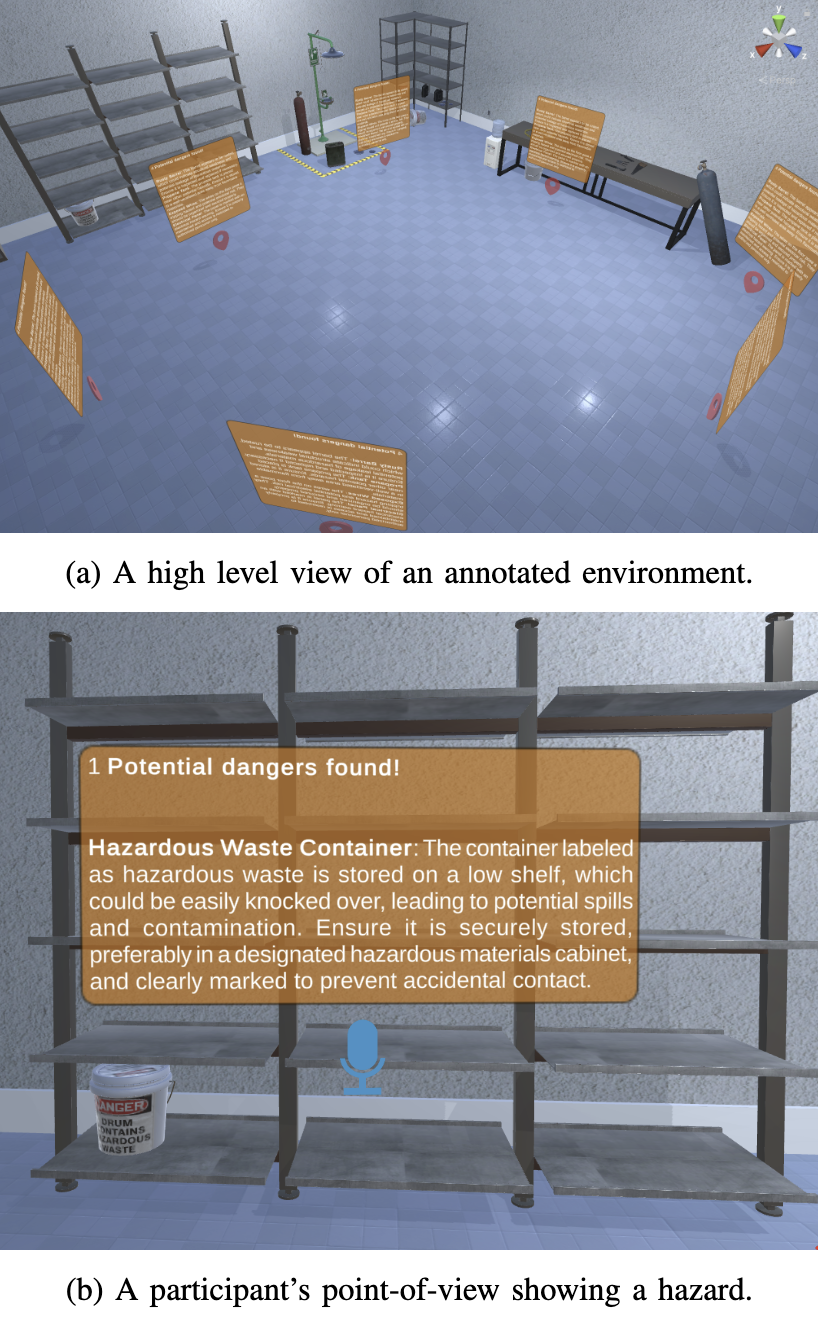}
    \caption{An example of a hazardous environment used in our experiments (a makerspace after an earthquake), with possible hazards autonomously annotated by interaction with a large vision-and-language model.}
    \label{fig:full_annotation}
\end{figure}

\begin{figure*}[t]
    \centering
    \includegraphics[width=0.6\textwidth]{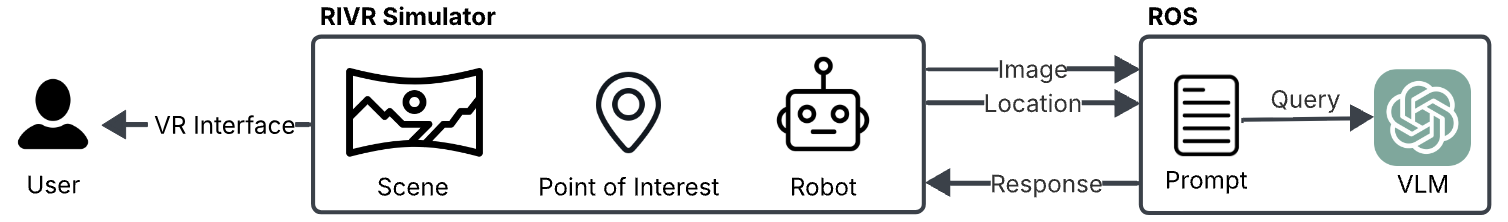}
    \caption{Our system pipeline for increasing situational awareness. The robot’s location and view data from the RIVR simulator are transmitted via ROS to a Vision--Language Model (VLM) for hazard assessment. The model’s response is returned to the user interface as identified points of interest.}
    \label{fig:system_overview}
\end{figure*}






\section{Introduction}
The 2023 train derailment in East Palestine released hazardous chemicals and forced large-scale evacuations, leading to a significant environmental and public health crisis. The investigations~\cite{reportPalestineb2023} show that failing to monitor potential hazards and maintaining situational awareness of emerging risks can cause minor hazards to escalate into  disasters~\cite{situatuinal_awareness_2008}. To prevent such large-scale failures, domains ranging from space exploration to search and rescue and industrial safety operate under strict standards, such as those established by the Occupational Safety and Health Administration (OSHA). These standards treat situational awareness as central to decision-making in high-risk environments. Yet even with established safety protocols, human operators struggle to maintain safety~\cite{misra2020informationOverload}. For example, fatigue in workspaces can cause hazards such as loose cables, spills, or obstructed exits to go unnoticed and lead to injury, while information overload can obscure critical cues. Maintaining situational awareness is critical not only for preventing accidents but also for supporting human decision-making during emergencies, such as when responders must rapidly identify hazards, prioritize risks, and coordinate actions under uncertainty. 
Autonomous robots that can provide situational awareness cues therefore offer a way to support operators in high-risk environments~\cite{industry4, understandingHRIinIndustry}, provided they are paired with effective Human–Robot Interaction (HRI).

In these settings, HRI increasingly incorporates Vision-Language Models (VLMs) because they combine visual perception with contextual reasoning~\cite{VLMsSurvey}. Their ability to interpret complex scenes could enable robots to meaningfully assist first responders and other safety-critical personnel. To explore this capability, we develop a VR-based framework that brings robotic perception and information presented to human operators into a controlled simulation environment for VLM-supported hazard annotation. In this paper, we describe a tool intended to reduce these risks by improving hazard awareness and risk management~\cite{DIQduringDisaster}. We imagine this tool to be used as part of a scenario in which an autonomous robot such as a drone is released to build up a virtual map of a high risk environment. Our system autonomously labels hazards in that environment, allowing a responder to explore the annotated environment in virtual reality.
\Cref{fig:full_annotation} shows an example of an annotated environment. 
In our approach, we capture images of the virtual-reality environment and present them to a large vision-and-language model (VLM), with instructions to identify any hazards present. The environment is annotated with the responses in the form of pop-ups that appear when a person exploring the space approaches a hazard. We refer to these annotations as points of interest (POIs) (\Cref{fig:full_annotation}). Our specific contributions are as follows:
\begin{enumerate}
    \item We present an open-source annotation system designed to increase situational awareness in virtual high-risk environments using VLM-based robots in the RIVR simulator~\cite{RIVR}.
    \item We present and evaluate an immersive interface featuring virtual markers with auditory message playback.
    \item Through a user study, we show that participants reported greater comfort and improved situational awareness when interacting with our system.
\end{enumerate}

\section{Background}
\label{sec:background}

\textbf{Risk management} is typically broken down into mitigation, preparedness, response, and recovery~\cite{DIQduringDisaster}. As preventing people from unexpectedly encountering hazards is ultimately more effective than reacting to them, our focus is on the proactive side of risk management, mitigation and preparedness. At the same time, the sometimes overwhelming flood of information during disaster response~\cite{DIQduringDisaster} means that even reactive strategies may benefit from structured hazard annotation. Our first hypothesis \textbf{(H1)} is that our VLM-based robot will successfully detect the majority (80\%) of hazards, while \textbf{(H2)} the interface will support people's situational awareness in dangerous situations, allowing information gathered by a robot to be explored safely.

\textbf{VLMs.} Despite growing interest in robotics for emergency response, industrial safety, and military operations, human-robot teams still struggle with adaptability, coordination, and clear communication~\cite{HRIinFutureMilitaryOperations2016, hoque2024hri}. One approach to managing human-robot interfaces is using natural language to enable robots to comprehend complex, context-rich commands so that non-expert users have an intuitive way of accessing information and interacting with a robot teammate~\cite{LLMs4HRI}. Recent advances in Large Language Models (LLMs) have been widely used to help robots understand~\cite{zhang2023LLMsEnhancingRobotUnderstanding} and plan over~\cite{joublin2023copalplanningrobot} human-like language. Similarly, large Vision-and-Language Models (VLMs) are particularly capable in visual reasoning tasks, and are currently being studied for purposes such as safety planning, navigation, and scene understanding~\cite{VLMsSurvey}.

There are a number of high-level ways in which LLMs and VLMs can be tied into HRI systems. They can be ``Scarecrows''~\cite{scarecrows} which are stand-ins for more principled approaches that will ultimately be replaced, or they can be a carefully selected component of a larger system in their own right. Our use of VLMs is in the latter category; we use careful prompt design and limited natural language interactions to construct a system that makes use of the power of large pre-trained models without the problems associated with direct human/model communication. Unlike studies that only discuss the interaction between humans and AI-enabled systems~\cite{setare2024operator5, scarecrows, modelingHRISystems}, we design and test this interface for a possible setting, and incorporate user feedback to refine it for practical future use.

\textbf{Simulation. }Virtual and augmented reality are progressively more integral to HRI research, and well-designed interfaces for them improve interactions between intelligent systems and operators~\cite{setare2024operator5}. We use the Robot Interactions in Virtual Reality (RIVR) simulator~\cite{RIVR}, a simulation environment designed to support Human-Robot Interaction in VR. RIVR's combination of VR interaction support, ROS robot management, and Unity environments allows us to implement an immersive user-friendly system of annotations for an intuitive user experience.

\textbf{Gaussian Splatting.} Creating realistic simulated environments (called digital twins) for human–robot interaction studies is often time-consuming and requires manual modeling in game engines. Recent advances in 3D scene reconstruction, particularly 3D Gaussian Splatting (3DGS), enable photorealistic environments to be generated directly from collections of images while maintaining real-time rendering performance. For example, aerial footage of the 2023 East Palestine train derailment was widely broadcast to provide situational updates. Such imagery was used in one of our scenes to reconstruct a photorealistic 3D representation of the environment using 3D Gaussian Splatting~\cite{shawn_gaussian}. These representations allow complex scenes to be reconstructed quickly and explored interactively, making them well suited for generating realistic environments for HRI experiments. 

\textbf{Hypotheses.} Building on these foundations, we hypothesize that VLM-supported robots can identify at least 80\% of hazardous items in a given space \textbf{(H1)}. Furthermore, we expect that people will generally feel improved situational awareness when navigating a scene with annotations as opposed to an un-annotated scene \textbf{(H2)}. 

To investigate these hypotheses, we present a VLM-powered system that autonomously annotates hazards in VR, aiming to improve robotic hazard communication, reduce operator cognitive overload, and enhance proactive risk management in HRI. We evaluate this system in a user study to assess its effectiveness in improving hazard perception and user comfort in VR-based HRI scenarios.

\section{Related Work}

Work has been conducted to explore the reliability of Vision-Linguistic Models (VLMs) during safety-critical recognition tasks. For example, Choi et al.~\cite{bettersafethansorry} describe the ``overreaction problem,'' where VLMs tend to over-react to safe situations and predict them as hazardous.

An important usage of VLMs is  context reasoning and anomaly detection in multimodal systems. For instance, Zhu et al.~\cite{ZhuLLMUnderstandVisualAnoms} explore the capability of large multimodal models to detect visual anomalies in a zero-shot setting, and Yang et al.~\cite{yang2025coinco} evaluate a model's ability to recognize objects in context from objects that are not. Compositional benchmarks, as presented in Winoground~\cite{winoground_2022}, focus on the assessment of reasoning failure in vision-linguistic models and contextual grounding. However, these studies focus more on contextual deviation rather than danger. Broader multimodal safety benchmarks~\cite{rottgerMSTSMultimodalSafety2025} assess unsafe or harmful multimodal model outputs; these works mainly focus on the generation of harmful advice or the execution of unsafe instructions, but not the classification of unsafe scenes.


VR can perform as an effective medium for hazard analysis and training in dangerous scenarios compared to conventional 2D methods like images on paper on presentations~\cite{chen_virtual_2018, toyodaVRbasedHealthSafety2022, slezakVirtualRealityApplication2018, puschmannRiskAnalysisAssessment2016}. Simulators already exist for harsh environments like space, but often do not have a good interface for human interaction; VR can be used to make training in such environments more stimulating~\cite{chen_virtual_2018}. Generalizing to other industries like mining, electronics, and construction, VR can be used to increase the immersion and presence of trainees in managing hazards; it is more immersive than CAVE systems or desktop VR, making the medium an effective tool~\cite{slezakVirtualRealityApplication2018, toyodaVRbasedHealthSafety2022}. Training presented in this manner has shown to be retained better over a period of several weeks~\cite{toyodaVRbasedHealthSafety2022}. Although this work does not directly address training, this existing work demonstrates the potential of VR as an effective interface for understanding risks.

In addition to training, VR has also proven effective in aiding hazard analysis. Simulations often require simplification of models at the risk of omitting information regarding hazards~\cite{puschmannRiskAnalysisAssessment2016}. Increasing the fidelity of a simulation and including human models increase the number of risks identified by participants~\cite{puschmannRiskAnalysisAssessment2016}. The ability for participants to perceive the true scale of objects may also aid in this process~\cite{slezakVirtualRealityApplication2018,puschmannRiskAnalysisAssessment2016}. However, it should be noted that the effectiveness of VR can be limited by the experience of the participants~\cite{slezakVirtualRealityApplication2018}.

Preceding work has also examined VR as an interface for presenting data from robotic exploration of a hazardous scenario~\cite{simmons2023hear}. It was shown that auditory cues can be generated by robots to help participants recognize hazards in a simulated scenario. However, a large number of sounds occurring at the same time proved stressful. Our approach offloads the duty of hazard recognition to VLMs; we hope this is a more comfortable user experience. 

\textbf{Gaussian Splatting.} Recent works~\cite{kerbl2023gaussian, shawn_gaussian} have explored neural scene representations such as Neural Radiance Fields and 3D Gaussian Splatting for reconstructing environments from images. Compared to NeRF methods, 3D Gaussian Splatting provides significantly faster rendering and supports real-time visualization of reconstructed scenes. These properties make it particularly suitable for interactive human–robot systems where operators must explore environments from multiple viewpoints.


\section{Methodology}
We implement a VR-based HRI system (\cref{fig:system_overview}) that integrates the RIVR simulator with ROS to support VLM-driven hazard annotation and real-time safety feedback. The interface presents identified hazards as interactive points of interest within the environment, allowing users to inspect, navigate, and receive concise explanations through visual markers and optional audio playback. The system is developed in Unity using the High-Definition Rendering Pipeline.

\subsection{System Design}

To simulate scenarios, various high-risk environments were considered, including a train derailment, a post-earthquake makerspace, and a construction site equipment failure. The makerspace setting was chosen for our user-study because the indoor setting facilitates navigating the whole environment in the virtual world, and offers a diverse range of potential hazards (hazardous chemicals, pressurized gas, tools, obstructed eyewash and fire suppressant stations). We simulated robot exploration of the environment with a Husky UGV with a full suite of sensors.  
During the robot's exploration of the room, it captures photos at a fixed rate and makes queries to the VLM service about safety issues in the image with an engineered prompt. The 19 ground-truth hazards were established by authors surveying the makerspace against OSHA's General Industry standards (29 CFR 1910)~\cite{osha_pha}.

After the robot has made a complete circuit of the environment and sent images to the VLM server, virtual Points of Interest are placed into the environment at the location of images with potential hazard. We define POIs as virtual markers that draw users' attention to certain areas in the simulator. These markers have two states: idle and activated. When idle, their visibility is a transparent red pin bouncing up and down; once a person enters the proximity of a marker (1 meter), it is activated, which brings up an information panel designed to present information in an intuitive way to help decision making. Pilot studies on the system suggested that some users found it difficult to read in virtual reality, motivating us to add a virtual button on the panel that allows users to listen to the annotations being read out loud. To do this, we used the audio file provided by ChatGPT-4o, generated separately for each response. 

For the implemented system, we used OpenAI's GPT, specifically model gpt-4o-2024-11-20. No fine-tuning was performed, and no seeds were specified. This model was selected based on its accessibility and performance on a range of queries obtained during pilot studies. We performed all queries during October and November 2024. To validate our model choice, we compared against Qwen2.5-VL-72B-Instruct, an open-weight alternative, on the same keyframes with the same prompt. GPT-4o outperformed it on every metric (GPT-4o: 94.7\% recall, 81.8\% precision, F1 = 0.878; Qwen: 31.6\% recall, 50.0\% precision, F1 = 0.39).

Currently, most of our scenarios are reconstructed manually using standard 3D modeling tools such as Blender and Unity3D, but future versions of the framework will mostly leverage scenes generated with 3D Gaussian Splatting, which enables faster and higher-quality environment reconstruction. As a proof-of-concept, we used drone footage of the train-derailment disaster in East Palestine to generate a 3D reconstruction of the scene using this approach; however, this environment was not included in the user study since it was too large to be fully covered by participants.

In future work, we plan to use this method to reconstruct photorealistic environments directly from RGB images and render novel viewpoints in real time, enabling rapid generation of interactive scenes for HRI experiments.


\subsection{Prompt}
\label{sec:prompt}

The model performs unified encoding of visual and textual inputs internally. This formatting ensures that the information is structured cleanly and enables the model to attend to the salient elements of the scene during inference. 

The visual input consists of RGB images
from our various environments and settings. These images
provide contextual views of critical equipment, materials, and
objects in the workspace and outdoors. All images are resized to 640×480
resolution before being passed into the model to match input
constraints and maintain visual consistency.

To keep our system generalizable to other environments with different safety concerns, we avoid specific details and scenario-dependent variables. Thus, we keep it simple by stating the room is a high-risk environment with potential hazardous items and situations. The main linguistic components of the prompt for the VLM, as used for the user study.

\begin{mdframed}[style=spacedframe,frametitle={Prompt}]
    \small
        You are an expert safety inspector specializing in hazard identification. Your primary role is to accurately and thoroughly identify potentially hazardous items in various environments. Provide clear, detailed explanations for why each item might be hazardous and suggest appropriate safety measures, ensuring high standards of safety are maintained. Respond concisely. Respond in the form of a JSON object with no extra words. Follow the given instructions to populate it.
        \begin{enumerate}
            \item Populate \texttt{hazards\_found} with the count of hazards found.
            \item Populate \texttt{safetyMessage} with hazard descriptions, starting with each hazard's name, followed by a caution note explaining why it is dangerous and what safety measures should be taken. Begin each hazard on a new line using \texttt{\textbackslash n} to create the line break.
            \item Populate \texttt{coordinates} with the location provided in \texttt{\{data.position\}}.
            \item Populate \texttt{image} with the path \texttt{\{data.image\}}.
            \item If at least one hazard is identified, set \texttt{generatePOI} to \texttt{true}, otherwise \texttt{false}.
        \end{enumerate}
    \normalsize
\end{mdframed}

\begin{mdframed}[style=spacedframe,frametitle={Response Example}]
    \small
    \begin{Verbatim}
{   "hazards_found": 1,
    "safetyMessage": 
    "1 Potential dangers found!
    Hazardous Waste Container: The
    container labeled as hazardous 
    waste is stored on a low shelf.",
    "image": "image.png",
    "generatePOI": true }
        \end{Verbatim}
    \normalsize
\end{mdframed}

\section{Experiments}

To evaluate the system, we conducted an IRB-approved user study using survey-based measures of usability and perceived situational awareness. 
We recruited 27 participants (ages 22–50) from a university setting with varied prior VR experience. All 27 participants experienced both conditions in a fixed order: they first explored an unannotated baseline scene, followed by exploration of the fully annotated VLM-assisted version of the same makerspace environment using a VR headset. Participants rated their experience on a 5-point Likert scale across five dimensions:
\begin{enumerate}
    \item \textbf{Clarity}: How clearly were hazards marked in the virtual environment?
    \item \textbf{Effectiveness}: How well did the system convey safety information?
    \item \textbf{Trust}: How confident are you in the system’s ability to detect hazards?
    \item \textbf{Usefulness}: How useful is the VR interface for improving safety awareness?
    \item \textbf{Future Use}: How likely are you to use this system in real-world scenarios?
\end{enumerate}

\begin{figure}
    \centering
    \includegraphics[width=\linewidth]{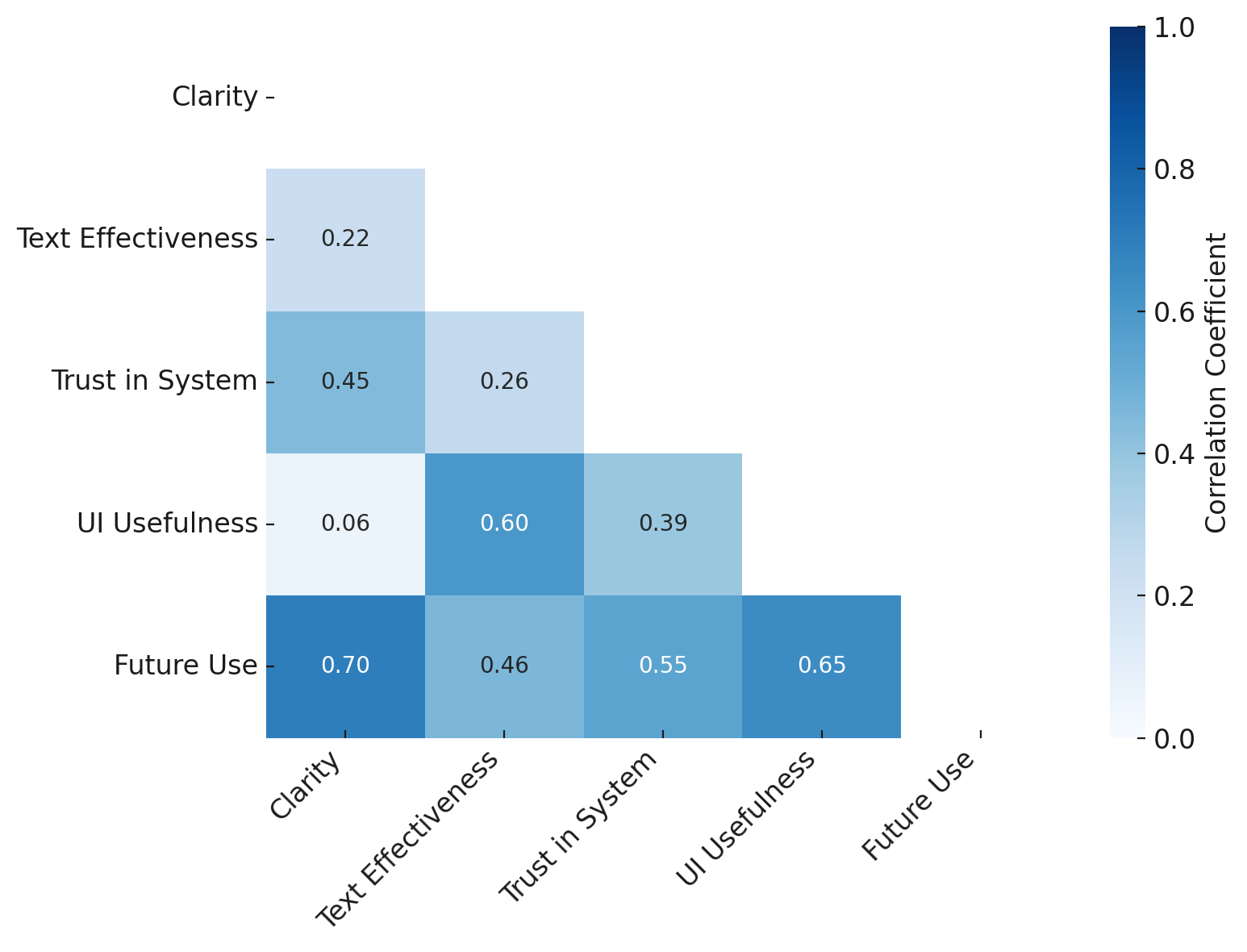}
    \caption{The correlation matrix shows correlations between the different components of the HRI system based on human feedback and suggests relatively high relationships across all five metrics (as expected given the system’s consistently high scores). For example, the strongest correlation (0.70) is between Future Use and Clarity, suggesting that participants who found the hazards easier to spot were more likely to adopt the system in the future.}
    \label{fig:ui_matrix}
\end{figure}

The correlation matrix (\cref{fig:ui_matrix}) highlights key relationships between system components and user perceptions. The strongest correlation is between Clarity of hazard indicators and Future Use (0.7), suggesting that the easier it is to spot hazards, the more likely users are to adopt the system. A similarly high correlation exists between UI Usefulness and Future Use (0.65), indicating that a well-designed interface also drives continued use. Trust in the system (0.55) is a factor, but not as strong as clarity or usability. Meanwhile, the low correlation between Clarity and UI Usefulness (0.06) shows that users view the clarity of hazard markers and the overall interface design as separate issues. This shows the importance of trust, clarity, and usefulness in letting non-experts use such systems effectively.

\pseudosection{Results and Analysis} We first review our initial finding from our user study for the VR interface evaluating the practicality of using VLMs for the hazard annotation task.

The average ratings for each of the five key aspects were consistently high, reflecting the system's effective performance, clear annotations, and strong user engagement. Clarity received the highest mean rating (4.6, SD = 0.65), indicating that most participants found the hazard indicators clear. Text Effectiveness was rated 4.25 (SD = 0.62), with suggestions for improved positioning and clarity. Trust in the System scored 4.16 (SD = 0.83), reflecting moderate trust, with real-world application identified as critical for confidence building. UI Usefulness (4.58, SD = 0.79) and Future Use (4.5, SD = 0.79) highlight the system's potential for operational adoption (see \cref{tab:wilcoxon}). Wilcoxon signed-rank tests confirmed that all five dimensions were rated significantly above the neutral midpoint of 3 (all $W \geq 210$, $p < .001$), indicating that participants' positive evaluations are statistically robust.

\begin{table}[h]

\centering
\caption{Wilcoxon Signed-Rank Test Results ($N = 27$, $H_1$: median $> 3$). All dimensions rated significantly above the neutral midpoint; $^{***}p < .001$, one-sided.}
\label{tab:wilcoxon}
\resizebox{\columnwidth}{!}{%
\begin{tabular}{lrrrrrr}
\toprule
\textbf{Dimension} & \textbf{Mean} & \textbf{SD} & \textbf{Median} & \textbf{W} & \textbf{\textit{z}} & \textbf{\textit{p}} \\
\midrule
Clarity              & 4.60 & 0.65 & 5.0 & 325.0 & 4.72 & $< .001$ \\
Effectiveness & 4.25 & 0.62 & 4.0 & 300.0 & 4.44 & $< .001$ \\
Trust                & 4.16 & 0.83 & 4.0 & 210.0 & 4.04 & $< .001$ \\
UI Usefulness        & 4.58 & 0.79 & 5.0 & 253.0 & 4.46 & $< .001$ \\
Future Use           & 4.50 & 0.79 & 5.0 & 250.0 & 4.25 & $< .001$ \\
\bottomrule
\end{tabular}%
}

\end{table}

Additionally, participants were asked to select whether they \textbf{preferred the annotated or unannotated environment} for safety assessment. Only one participant preferred the unannotated scene over the annotated one. Finally, an \textbf{open-ended response section} allowed them to provide \textbf{comments and suggestions} for system improvements. This helped us to better understand our system's shortcomings. Most people showed much enthusiasm towards our system, working with robots, and VR. Some suggested that the information panel’s height should be user-adjustable for easier reading. Additionally, they recommended adding a minimize feature to allow users to hide or show the panel as needed. A few recommended integration of other sensors like temperature, pressure, and sound. One suggested to use more recent VR devices such as Quest 3 for a wireless experience. 
The user study highlighted that VLMs show promise for automatically identifying hazards and communicating safety-relevant information in simulated environments.
The human feedback from our user study strongly supports \textbf{H2} that people generally feel more aware and at ease navigating a scene with annotations as opposed to one without any. Wilcoxon signed-rank tests ($N = 27$) confirmed that ratings for all five dimensions were significantly above the scale midpoint (all $p < .001$), indicating that users significantly trust and accept LLM-supported robots for safety awareness.

We have demonstrated how capabilities of GPT-4o can be utilized in robot-assisted hazard detection, enabling them to perform as a sociable, useful assistant in critical situations. Out of the 19 hazards placed at this scenario, the system missed only one potential hazard and successfully identified the rest. This supports \textbf{H1}: VLMs are able to identify most (more than 80\%) of the hazards in the space. Of the 19 hazards present in the environment, GPT-4o produced 4 false positives (21.1\%) and 1 false negative (5.3\%). The false positives fell into two categories. Two were outright hallucinations; for instance, in a scene containing only an unsecured gas cylinder and a blocked safety shower, the model additionally reported a flammable-liquid container that was not present in the image. The other two were mislabelings of real hazards; the same gas cylinder, for example, was tagged as ``in a safety zone'' rather than ``unsecured.'' The single false negative was a cylinder occluded behind another object. Notably, all failure cases occurred in scenes where objects were in close proximity, which suggests that denser images will increase error rates.


\section{VLM Reporting and Ethical Considerations}

The use of large pre-trained models has, or has the potential to have, significant implications regarding privacy, social justice, environmental concerns, and reproducibility. In this section we briefly describe the specifics of the model used and the impact of our work.

\textbf{Privacy:} Although we let humans virtually interact with the scene, no human avatar is included in GPT's input images for the Sim2Real process in this study due to privacy and safety concerns. Additionally, human participants are not interacting directly with the VLM, reducing the risk of data leakage based on human inputs.

Future studies and applications will likely incorporate human avatars resembling study participants. Human state is important to the task and could lead to life-threatening consequences if ignored; however, sending images of people to GPT may not be in their best interest in regards to their privacy. Another concern is protecting sensitive information in the environment itself, such as proprietary technology. A potential solution to this is running models locally on the robot; however, this may undermine the performance requirements for accurate hazard annotation, exacerbating accuracy issues. Therefore, more research such as~\cite{keepSecret} that evaluates how much VLMs are able to keep secrets is essential. Work on smaller multimodal language models like TinyLLaVa~\cite{zhouTinyLLaVAFrameworkSmallscale2024} may also enable all computation to be done locally, limiting the scope information is shared in.

\textbf{Social implications:} Because the VLM is not interacting directly with human participants, concerns such as models that perform unevenly based on voice or skin tone~\cite{lee2024skincolor,birhane2024dark} are not relevant to this study, nor is demographic information (or any user specific information) transmitted to the model.

\textbf{Hallucinations:} To maximize situational awareness, humans need to rely on the information being presented in the environment. Supporting human trust crucially requires that users be able to observe the system's reasoning in identifying certain items as hazardous, which is currently an opaque operation. To address this concern, in future we will explore using Chain of Thought (CoT)~\cite{cot} and other XAI~\cite{xai} techniques. This would be beneficial in cases where similar hazards need to be treated differently in different environments, so as to ensure that the system performs equitably across all environments and scenarios by conducting a series of various experiments in different spaces. Integrating human feedback to fine-tune the VLMs and the pipeline as a whole may help avoid false positives, negatives, and hallucinations in different environments~\cite{yuanTrustworthyAnomalyDetection2022}. 

\textbf{Environmental:} Environmentally, although detailed figures are difficult to obtain, a conservative estimate is that each of our GPT inferences is responsible for approximately 0.047 kilowatt-hours of electricity~\cite{luccioni2024power}. We estimate that the electricity usage of our study was roughly 1.88 kWh (0.047 times the approximately 40 images evaluated). This does not include one-time development costs. Novel CoT models like DeepSeek R1~\cite{deepseek2025deepseekr1} may help to address environmental concerns by reasoning through what safety standards are relevant for a given scenario with less computational power.

\textbf{Omitting the VLM:} Without the use of a vision and language model, this work would have relied on simpler methods of automatically annotating points of interest, possibly with manual involvement. However, because the VLM is core to the contributions of this work, ablation studies were not performed.

\section{Limitations}

One limitation of the work to date is the size of our user study, which contained 27 participants. Although it is our intention to conduct larger studies in future, it is also possible that future large-scale evaluations could utilize platforms such as Prolific to allow more people to experiment with the interface. This is made more difficult by the requirement that participants have access to a VR headset, but as such equipment becomes more widespread, crowdsourcing VR work become a more realistic possibility.

Our evaluation relies primarily on a single VLM, which provides a stable testing environment but limits the scope of the findings. Recent vision–language models, both large and smaller, offer diverse architectural biases and training methods. In future we will compare our system against additional baselines in order to determine comparative performance.

\section{Conclusion and Future Work}

In this paper, we presented and evaluated a framework that automatically annotates potential risks to support users’ situational awareness in hazardous environments. We have observed promising results of VLMs automatically annotating potential dangers; practically, such a tool has the potential to be useful in disaster settings such as our earthquake scenario, or in inherently risk-filled settings such as construction sites. This study supports the cautious integration of large pretrained models in HRI systems~\cite{kim2024understanding} as a source of environmental analysis for applicable scenarios. Despite a limited sample size, the results suggest that users found the system satisfactory. Assistant robots can play a meaningful role in enhancing safety awareness in high-risk environments, such as disaster response and industrial settings. 

There are many avenues for future work that build upon the foundation of the methodology presented in this paper, as well as the strong findings of the user study for utilizing VLMs for effective hazard analysis in the VR interface with POI indicators.
As methodological enhancements, prior work has shown that imposing a coordinate grid on image inputs for VLMs can significantly improve question-answering, image captioning, and segmentation~\cite{zhangAgent3DZeroAgentZeroshot2024}. Future work may improve our system by modifying input images with such a grid or other tools to help the VLM better understand the scene. Since VLMs are starting to support video analysis, future work should also explore whether providing a video (rather than downsampled still images from the robot's video camera) increases annotation accuracy.

Future work could also explore active hazard response in VR, such as virtually resolving problems with a robot (e.g., straightening a precariously balanced waste container) to simulate mitigation. Task prioritization and resource management in such hazard response scenarios could also be investigated. The scope of this study should also be expanded to include evaluation by personnel such as first responders.

Disaster scenarios are also dynamic; annotations produced may rapidly become outdated. Future research should explore how autonomous agents can adapt to these changing environments, such as tracking the spread of fire or monitoring the status of injured victims. Modeling these evolving conditions in our simulation framework could allow us to test dynamic analysis informed by hazard message reports. These results should also be evaluated in real-world conditions through the Sim2Real~\cite{simulationsInRobotics} process before practical use. We advocate for the use of such digital twins, as they can represent the status, features, and behavior of their physical twins in real time with high accuracy~\cite{digitalTwin_Han_2022}.

\bibliographystyle{IEEEtranS}
\bibliography{new_references}


\end{document}